%% file: main.tex
\documentclass[10pt,twocolumn,letterpaper]{article}

\usepackage[pagenumbers]{cvpr} 
\usepackage{pifont}
\usepackage{xcolor}

\input{preamble}

\definecolor{cvprblue}{rgb}{0.21,0.49,0.74}
\usepackage[pagebackref,breaklinks,colorlinks,allcolors=cvprblue]{hyperref}

\def\paperID{*****} 
\def\confName{CVPR}
\def\confYear{2025}

\title{Poutine: Vision-Language-Trajectory Pre-Training and Reinforcement Learning Post-Training Enable Robust End-to-End Autonomous Driving}

\author{Luke Rowe\thanks{Equal contribution.}$^{*1,2}$, Rodrigue de Schaetzen\footnotemark[1]$^{*1,2}$, Roger Girgis$^{1,3}$, Christopher Pal$^{1,2,3,4}$, Liam Paull$^{1,2,4}$ \\
$^1$Mila - Quebec AI Institute, $^2$Université de Montréal, $^3$Polytechnique Montréal,
$^4$CIFAR AI Chair\\
}

\begin{document}
\maketitle
\input{sec/0_abstract}

\input{sec/1_intro_report}
\input{sec/2_relatedwork}
\input{sec/3_methods}
\input{sec/4_results}
\input{sec/5_conclusion}
{
    \small
    \bibliographystyle{ieeenat_fullname}
    \bibliography{main}
}

\clearpage
\onecolumn
\appendix

\input{sec/X_suppl}  


\end{document}

%% file: preamble.tex
\usepackage{makecell}
\usepackage{listings}     
\usepackage{setspace}     
\lstdefinestyle{prompt}{
  basicstyle=\fontsize{6}{5.8}\selectfont\ttfamily,   
  breaklines=true,                   
  frame=single,                      
  backgroundcolor=\color{gray!5},    
  rulecolor=\color{gray!70},
  xleftmargin=0pt, xrightmargin=0pt,
  aboveskip=4pt, belowskip=4pt,
  breaklines=true,
  breakautoindent=false,   
  breakindent=0pt,         
  mathescape=true,
}

%% file: sec/0_abstract.tex
\begin{abstract}
Maintaining good driving behavior in out-of-distribution scenarios remains a critical challenge in autonomous driving. A promising direction is to leverage the generalist knowledge and reasoning capabilities of large-language models by treating unusual driving scenarios as a logical reasoning task. 
In this work, we present Poutine, a method that uses an off-the-shelf 3B-parameter vision-language model (VLM) — without any additional components — to achieve robust end-to-end autonomous driving via a simple and scalable training recipe.
To learn strong base driving capabilities, we first train Poutine-Base using self-supervised next-token prediction over vision, language, and trajectory (VLT) tokens, leveraging both nominal and long-tail driving data. In the second stage, we fine-tune Poutine-Base using Group Relative Policy Optimization (GRPO) with a small set of human preference-labeled examples. We evaluated our approach on the Waymo end-to-end driving benchmark curated for long-tail scenarios. The final Poutine model achieves an \textbf{RFS of 7.99} on the test set, placing 1st\footnote{Our submission topped the public leaderboard but, under the competition’s Quebec-residency exclusion, was not prize-eligible; the organizers instead granted us a \emph{special mention}.\\Challenge public leaderboard: \url{https://waymo.com/open/challenges/2025/e2e-driving/}} in the 2025 Waymo Vision-Based End-to-End Driving Challenge by a significant margin. Our results suggest that handcrafted tokenizers or custom architectural components added to base VLMs in prior work are not necessary to achieve strong driving performance. Instead, this work highlights the potential of scalable VLT pretraining combined with lightweight RL fine-tuning to enable robust and generalizable autonomous driving. 

\end{abstract}

%% file: sec/1_intro_report.tex
\section{Introduction}
\label{sec:intro}

Vision–language models (VLMs) have emerged as a powerful means of coupling visual perception with the world knowledge and common sense reasoning acquired from internet-scale pre-training \citep{bai2023qwen, gpt42023openai, chen2024internvl, dubey2024llama3, reid2024gemini}. For autonomous vehicles, such multimodal reasoning is most valuable in long-tail situations—rare but safety-critical events that dominate operational risk and have limited coverage in conventional driving corpora. Nevertheless, the empirical study of VLMs for driving has so far been restricted largely to nominal driving benchmarks such as nuScenes~\citep{caesar2020nuscenes}, where high-level semantic reasoning is seldom required and the benefits of language grounding remain unclear \citep{hedge2025dima, jiang2024senna, pan2024vlp, sima2024drivelm, nie2024reason2drive, bai2024atlas, tian2024drivevlm, tian2024token, xing2025openemma, huang2024rdadriver, wang2024omnidrive, zhou2025opendrivevla, mao2023gptdriver, hwang2024emma, xu2024vlmad}. The Waymo Vision-Based End-to-End Driving (WOD-E2E) dataset \cite{waymo2025e2e}, along with its accompanying challenge, presents a new opportunity for researchers to test trajectory planning in more challenging curated long-tail scenarios that account for less than 0.003\% of daily driving. This is in stark contrast to the relatively boring scenarios present in other open-loop planning benchmarks. Hence, the WOD-E2E dataset provides a more suitable testbed for investigating whether the generalist knowledge embedded in VLMs can translate into safer and more reliable driving policies.

To address this question, we introduce \textit{Poutine}\footnote{Named after the Québécois dish which combines fries, gravy, and cheese curds.}, a 3B-parameter VLM for end-to-end driving, trained via a simple two-stage pipeline that fine-tunes a pre-trained Qwen2.5-VL model~\cite{Qwen2.5-VL}. In Stage 1, \emph{Poutine-Base} learns base driving capabilities by performing standard next-token prediction over vision, language, and trajectory (VLT) input tokens to predict both a driving explanation and future trajectory, where the trajectories are represented as text. The training corpus comprises 83 hours of nominal Japanese driving from the public CoVLA dataset \citep{arai2025covla} and 11 hours of long-tail driving from WOD-E2E. All language annotations are generated automatically by a 72B-parameter VLM, removing the need for manual labeling and yielding a fully self-supervised pre-training procedure that is straightforward to scale. In Stage 2, we refine Poutine-Base with Group Relative Policy Optimization (GRPO) using less than 500 human preference-labeled frames from the WOD-E2E validation set. Despite the modest supervision budget, this lightweight reinforcement-learning (RL) step materially improves policy performance in the long-tail. 

The trained Poutine-Base model achieves a rater-feedback-score (RFS) of 8.12 on the WOD-E2E validation set, nearly matching the 8.13 RFS of Waymo’s expert ground-truth trajectories of the ego vehicle. Moreover, a variant trained solely on the CoVLA Japanese driving dataset generalizes zero-shot to US driving data, achieving an RFS of 7.74 on the WOD-E2E validation set despite never encountering Waymo data during pre-training. This result highlights the potential of collecting diverse driving data from various geographical regions to train a unified driving policy via large-scale next-token prediction VLT pre-training. After GRPO fine-tuning, the final Poutine model attains a 7.99 RFS on the official WOD-E2E test set, achieving first place in the 2025 Waymo Vision-Based End-to-End Driving Challenge by a large margin. Collectively, our results indicate that scalable VLT pre-training, coupled with lightweight preference-based RL, constitutes a practical recipe for robust and generalizable autonomy in challenging long-tail driving scenarios.

 \textbf{Contributions:} (1) We present Poutine, a simple and scalable VLM-based E2E driving policy that combines large-scale VLT pre-training with lightweight GRPO fine-tuning. We show that the combination of next-token prediction pre-training followed by RL finetuning that has proven successful in other domains \citep{ouyang2022rlhf, deepseek2025r1} can yield robust E2E motion planning policies. (2) We show that Poutine-Base enables zero-shot transfer to unseen geographic regions, as the model variant pre-trained exclusively on Japanese data generalizes zero-shot to the US despite the substantial distribution shift (i.e., left- versus right-hand driving). (3) Poutine secures first place in the 2025 Waymo Vision-Based End-to-End Driving Challenge, establishing a new benchmark for long-tail autonomy.

%% file: sec/2_relatedwork.tex
\section{Related Work}
\label{sec:relatedwork}

\textbf{VLMs for End-to-End Driving.} Several prior works finetune VLMs for end-to-end driving in the open-loop \citep{sima2024drivelm, hwang2024emma, wang2024omnidrive, tian2024token, bai2024atlas, huang2024rdadriver, zhou2025opendrivevla, xu2024drivegpt4} and closed-loop \citep{shao2024lmdrive, renz2025simlingo, zhang2024wisead, wang2023drivemlm, fu2025orion, paul2024legodrive} setting. In both settings, evaluations are performed on nominal benchmarks such as nuScenes or on synthetic driving benchmarks like CARLA whose scenario diversity is limited; none report results on a curated real-world long-tail corpus comparable to WOD-E2E, which is the focus of the present work. A separate line of research employs VLMs as auxiliary components—either as natural-language conditioners for conventional end-to-end stacks \citep{jiang2024senna, tian2024drivevlm} or as teachers whose features are distilled into end-to-end stacks \citep{xu2024vlmad, hedge2025dima, pan2024vlp}. Poutine is a VLM finetuned directly for end-to-end driving, but unlike prior work, we additionally leverage RL post-training to enhance the model's driving capabilities in long-tail scenarios. 

Prior works use custom perception backbones \citep{wang2024omnidrive, tian2024token, zhou2025opendrivevla} and action headers \citep{fu2025orion, renz2025simlingo, shao2024lmdrive} that are aligned to language space. To maintain the simplest design, we finetune an off-the-shelf VLM without any architectural modifications —leveraging both its vision encoder and language decoder—and predict the trajectory directly as natural language tokens. This alleviates the need for custom visual perception backbones or trajectory headers. The most comparable prior work is EMMA \citep{hwang2024emma}, which likewise fine-tunes an off-the-shelf VLM for end-to-end driving. In contrast to EMMA’s purely supervised fine-tuning approach, Poutine supplements supervised learning with preference-based RL, yielding improved driving capabilities in long-tail scenarios.
\\\\
\textbf{RL Post-Training for End-to-End Driving.} TrajHF \cite{li2025finetuning} employs preference-based reinforcement learning to align open-loop trajectory predictions with human preferences, but its fine-tuning stage consumes 78 000 labelled frames and is applied to a model that lacks large-scale VLM pre-training. In contrast, Poutine bootstraps from both internet-scale VLM pre-training and our vision–language–trajectory (VLT) pre-training, achieving more human-aligned planned trajectories with only 479 preference-labelled frames. Group Relative Policy Optimization (GRPO) has recently proved effective for boosting reasoning capabilities in language models \citep{deepseek2025r1} and has been adopted by AlphaDrive \citep{jiang2025alphadrive} to refine high-level natural-language behaviour planners. AlphaDrive, however, does not generate planned trajectories. By contrast, Poutine leverages GRPO to directly improve trajectory planning performance.  

Concurrent to our work are AutoVLA \cite{zhou2025autovla} and Drive-R1 \cite{li2025drive}, which also explore combining self-supervised pretraining with GRPO fine-tuning for VLM-based trajectory planning in autonomous driving. However, AutoVLA  reports substantially lower performance on the WOD-E2E benchmark, suggesting that its physical action token generation component may negatively impact overall performance. Meanwhile, Drive-R1 is only evaluated on nuScenes, a benchmark that consists primarily of nominal driving scenarios.

%% file: sec/3_methods.tex
\begin{figure*}[t]
     \centering
     \includegraphics[width=\linewidth]{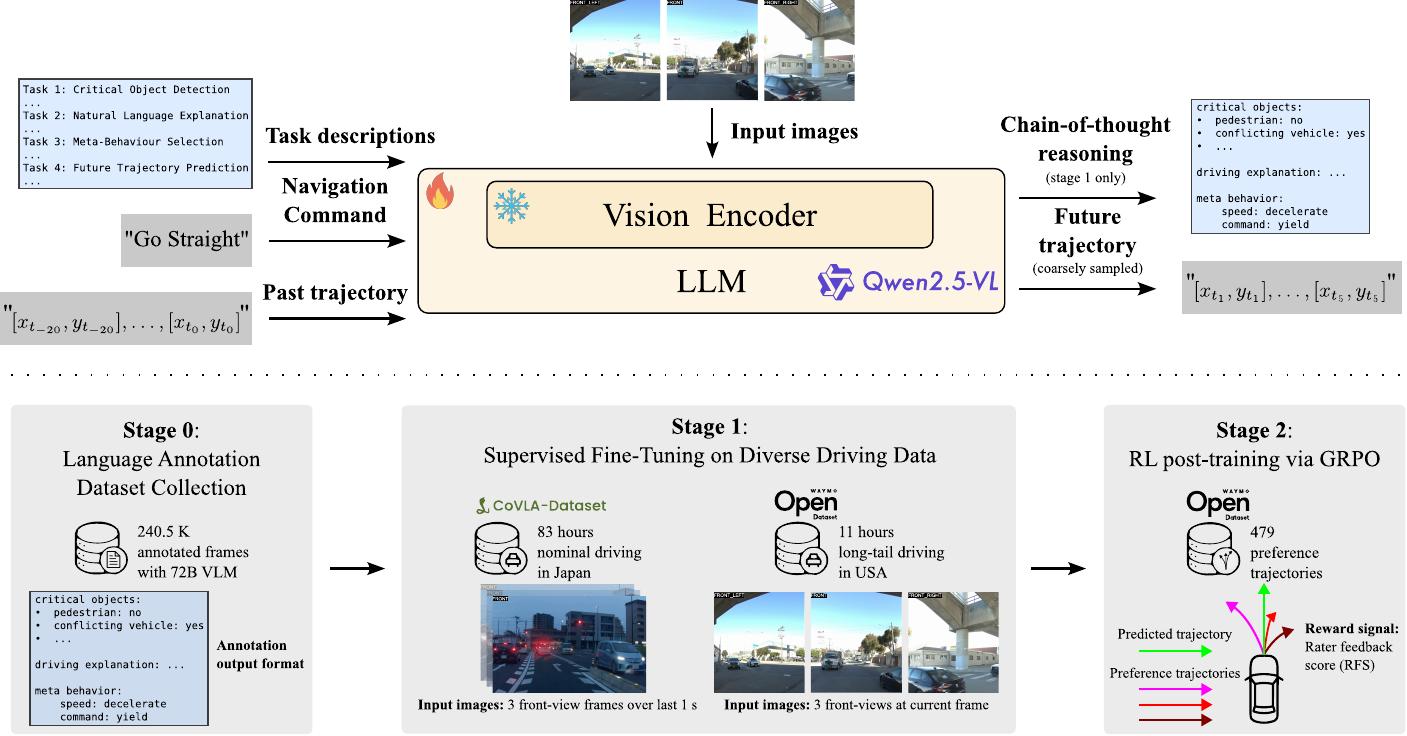}
     \caption{(Top row) The Poutine model adopts an off-the-shelf 3B-parameter VLM with no additional components, and simply encodes trajectories in text space. (Bottom row) The proposed pipeline to train an end-to-end planner for robust autonomous driving. Datasets shown in the figure (i.e., CoVLA \cite{arai2025covla} and WOD-E2E \cite{waymo2025e2e}) indicate those used in our experiments. Fine-tuning of the Poutine model follows the highlighted stages with the vision encoder kept frozen.}
     \label{fig:intro-fig}
 \end{figure*}

\section{Poutine}

Figure \ref{fig:intro-fig} provides an overview of our trajectory‑planning framework for autonomous driving. Our model builds on an autoregressive, transformer‑based VLM pre‑trained on internet‑scale text and image data. In this work, we adopt the \texttt{Qwen2.5-VL 3B Instruct} model \cite{Qwen2.5-VL}, which combines a Vision Transformer (ViT) encoder with a decoder‑only transformer language model. No additional architectural components or custom tokenizers are introduced to the Qwen model. As such, we treat trajectory planning as a next‑token prediction task over the original token vocabulary, thus representing trajectories directly as text.

The Poutine model ingests a task description, the current high-level navigation command (referred to as the \emph{intent}), the past ego trajectory, and a sequence of historical multi‑view RGB images captured by cameras mounted on the ego‑vehicle. We train our model in two output modes: 1) chain‑of‑thought (CoT) reasoning followed by the predicted future trajectory of the ego, and 2) the future trajectory only. Our proposed training pipeline consists of two stages. During the first stage, we perform supervised fine-tuning (SFT) on a large, curated driving dataset made up of both nominal and long‑tail scenarios. The ground truth future trajectories of the ego as well as automatically generated language annotations serve as the training signal for SFT. We refer to the trained model from Stage 1 as \emph{Poutine‑Base}. Following SFT, we fine-tune our Poutine-Base model using the proposed RL post-training stage leveraging the GRPO algorithm. These two training stages as well as our approach to automatically collecting language annotations are described next.

\subsection{Language Annotation Dataset Collection}
\label{sec:ann}

\begin{figure*}[t]
    \centering
    \includegraphics[width=1\linewidth]{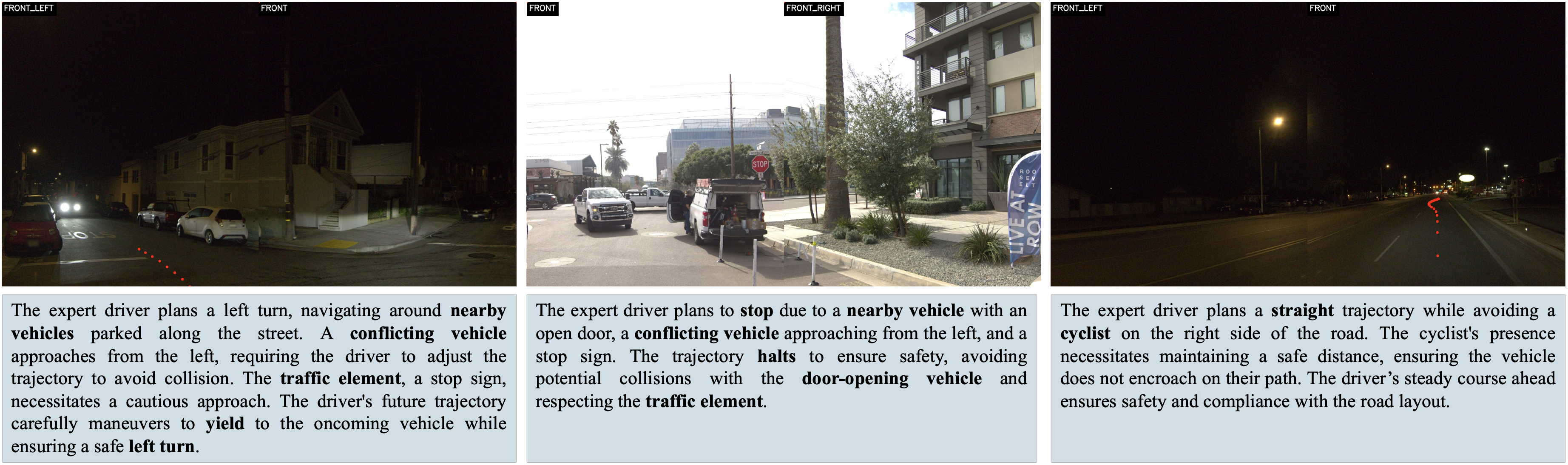}
    \caption{\textbf{Generated annotations on WOD-E2E data.} The red dots depict the 5-second future trajectory. Only two views from the current frame and the driving description of the annotation are shown. Bold text highlights the objects and meta behavior selected by the model.}
    \label{fig:annotation_example}
\end{figure*}

Incorporating language annotations as part of the training targets allows us to introduce an auxiliary scene‑understanding task to boost downstream trajectory planning performance. The goal is for the generated captions to supply trajectory‑specific context that the model can attend to during planning. Moreover, CoT reasoning offers a promising way to tackle out‑of‑distribution scenarios, where planning may be solved via logical reasoning.

We present our automated pipeline for generating language annotations on driving data, eliminating the need for manual labeling or post‑hoc verification. Annotations are generated in a zero‑shot manner using the \texttt{Qwen2.5‑VL 72B Instruct} model where we condition on the same set of inputs discussed in the previous section (see Figure \ref{fig:intro-fig}, top). In contrast to most driving datasets with language and action data, we also condition on the ground-truth future trajectory of the ego vehicle. The resulting captions therefore describe why the vehicle executes the given future trajectory rather than predict where the ego should go based on the past context. We found that this significantly improves the consistency between the language annotations and ground-truth future trajectories.

The language annotation task is divided into three components: 1) identify the relevant critical objects in the current scene from a predefined list, 2) compose a short description explaining why the expert trajectory was executed, referencing the selected critical objects, and 3) identify the meta planning behavior from a predefined list of possible speed and command actions. Preliminary testing showed that the annotation quality dramatically improves when the critical object detection task is structured as yes or no questions compared to more open-ended descriptions of the task. Several example annotations are shown in Figure \ref{fig:annotation_example} and the system prompt is provided in Figure \ref{fig:system-prompt-ann} in Appendix \ref{app:annotation}.

\subsection{Vision-Language-Trajectory Pre-Training}





Several design decisions were made to facilitate alignment between the image, language, and trajectory modalities during the SFT stage. To simplify training, the model predicts the future trajectory at a low temporal resolution ($\Delta t = 1$~s in our experiments) and later upsample, if necessary, using cubic spline interpolation. To discourage shortcut learning, we remove the past trajectory and intent during an initial phase of VLT pre-training, forcing the model to derive useful representations from the higher-dimensional image–caption signal rather than relying on the lower-dimensional cues.

The collected annotation dataset (Section \ref{sec:ann}) allows us to recast trajectory planning as a four-stage chain-of-thought (CoT) reasoning sequence: (i) detection of critical objects and conditions, (ii) generation of a natural-language explanation, (iii) meta-behaviour selection, and (iv) prediction of the future path. The model executes four successive question–answer turns, each stage conditioning on the previous response. Figure \ref{fig:system-prompt-sft} in Appendix \ref{app:annotation} shows our system prompt for VLT pre-training.

\subsection{Reinforcement Learning Post-Training}

We introduce a reinforcement learning post-training stage leveraging the GRPO algorithm \cite{shao2024deepseekmath} to fine-tune the Poutine-Base model on a set of high quality examples. In this stage, we fine-tune only on the future trajectory prediction task and omit outputting a structured CoT reasoning trace. This design decision significantly improves inference latency without impacting performance, as will be made clear in Section \ref{sec:results}. 

Let $\pi_{\theta}$ and $\pi_{\text{ref}}$ denote the optimized policy and the pre-RL policy (Poutine-Base), respectively. Given the policy $\pi_{\theta_{\text{old}}}$ from the previous iteration of GRPO, we sample $N$ outputs $\{o_i\}_{i=1}^N$ for the given prompt $q$. Each of these outputs from the model, which we expect to be a well formatted trajectory, is assigned a group-relative advantage $A_i$ representing the quality of the predicted trajectory according to the specified reward function $r$, relative to the other candidate predictions:
\begin{equation}
A_i = \frac{r_i - \text{mean}(\{r_j\}_{j=1}^N)}{\text{std}(\{r_j\}_{j=1}^N)}.
\end{equation}
Hence, the advantage is the normalized reward where we use the the mean reward, averaged across the sampled outputs, as the baseline. 

To incentivize the model to generate outputs with high advantage within its group, we update the policy $\pi_{\theta}$ according to the objective

\begin{align}
\mathcal{J}_{\text{GRPO}}(\theta) &= \mathbb{E} \left[\{o_i\}_{i=1}^N \sim
\pi_{\theta_{\text{old}}}(q)\right ] \\
& \frac{1}{N}\sum_{i=1}^N \{ \min [s_1 \cdot A_i, s_2 \cdot A_i] - \beta \mathbb{D}_{\text{KL}} [\pi_{\theta} \mid\mid \pi_{\text{ref}}] \} \notag\\
s_1 &= \frac{\pi_{\theta}(o_i\mid q)}{\pi_{\theta_{\text{old}}}(o_i \mid q)} \\
s_2 &= \text{clip}\left ( \frac{\pi_{\theta}(o_i\mid q)}{\pi_{\theta_{\text{old}}}(o_i \mid q)}, 1 + \epsilon, 1 - \epsilon \right ),
\end{align}
where hyperparameters $\epsilon$ and $\beta$ determine the clipping range and the weight of the KL divergence term $\mathbb{D}_{\text{KL}}$, respectively. The latter  ensures the policy does not deviate significantly from the original model pre-trained during the SFT stage. Note that in contrast to the previous stage, supervision occurs on a per full output basis (e.g., the entire trajectory in text space), not on a per token basis.

We use a reward function consisting of two terms: a driving specific reward $r_{\text{drive}}$ and a format reward $r_{\text{format}}$:
\begin{align}
r &= r_{\text{drive}} + r_{\text{format}}\\ & r_{\text{drive}} \in [0, 1], \ r_{\text{format}} \in \{0, 1\}
\end{align}
The driving reward can simply be the same objective used during SFT (i.e., the L2 error between the predicted and the ground truth future trajectory), or it can be derived using a more sophisticated training signal such as a preference score from a human annotated dataset. The format reward, on the other hand, gets assigned a value of 1 if the model outputs a correctly formatted trajectory and 0 otherwise.

%% file: sec/4_results.tex
\section{Results}
\label{sec:results}

\begin{table*}[t]
  \centering
  \caption{Overview of our language annotation dataset.}
  \label{tab:ann-dataset-stats}
  \small                    
  \begin{tabular}{lrrrrrr}
    \toprule
    Dataset & \makecell[r]{Annotated\\Scenarios} & \makecell[r]{Annotated\\Frames} & \makecell[r]{Annotation\\Frequency} &  \makecell[r]{Input frame window} & \makecell[r]{Camera\\Views} \\
    \midrule
    WOD-E2D (train)   &   2037  & 39.5K & 1 Hz & $t=[-1.0,\,-0.5,\,0]$ & 8  \\[2pt]
    CoVLA    &   10,000   & 201K & 1 Hz & $t=[-1.0,\,-0.5,\,0]$& 1  \\
    \bottomrule
  \end{tabular}
\end{table*}

\begin{table*}[t]                
  \centering
  \small                           
  \begin{tabular}{lcccc}
    \toprule
    Method &
    \textbf{RFS$^{\dagger}$ (Overall)} $\uparrow$ &
    \textbf{RFS (Spotlight)} $\uparrow$ &
    \textbf{ADE at 5 s (Avg)} $\downarrow$ & 
    \textbf{ADE at 3 s (Avg)} $\downarrow$
    \\
    \midrule
    Ego-status MLP & 7.41 & 6.19 & 2.94 & 1.28 \\
    \midrule
    Waymo Baseline & 7.53 & 6.60 & 3.02 & 1.32 \\
    Swin-Trajectory \cite{zhou2025swinfrajectory} & 7.54 & 6.68 & \underline{2.81} & \textbf{1.21} \\
    AutoVLA \cite{zhou2025autovla} & 7.56 & \underline{6.94} & 2.96 & 1.35 \\
    UniPlan \cite{liao2025diffusiondrive} (DiffusionDrive)  & 7.69 & 6.65 & 2.99 & 1.31 \\
    DiffusionLTF \cite{Nguyen2025DiffusionLTF} & 7.72 & 6.41 &  2.89 & 1.36 \\
    HMVLM \cite{wang2025hmvlm} & 7.74 & 6.73 & 3.07 & 1.33 \\
    ViT-Adapter-GRU & 7.85 & 6.67 & 2.89 & 1.44 \\
    \midrule
    Poutine-Base & \underline{7.91} & \textbf{7.04} & 2.94 & \underline{1.27} \\
    Poutine & \textbf{7.99} & 6.89 & \textbf{2.74} & \textbf{1.21} \\
    \bottomrule
  \end{tabular}
  \caption{\textbf{Results on WOD-E2E test split.} Best-performing model \textbf{bolded} and second-best \underline{underlined}. The results are taken from the public leaderboard where we only show methods that scored higher than the Waymo baseline.$^{\dagger}$ RFS range: 4–10.}
  \label{tab:waymo-test}
\end{table*}

\begin{table*}[t]                
  \centering
  \small                           
  \begin{tabular}{lcccccc}
    \toprule
    & \multicolumn{2}{c}{VLT Pre-Training} & \multicolumn{2}{c}{CoT} & & \\
    \cmidrule(r){2-3} \cmidrule(r){4-5}
    Method &
    CoVLA &
    WOD-E2E &
    Train &
    Test &
    \textbf{RFS (Overall)} $\uparrow$ \\
    \midrule
    Qwen 2.5-VL 3B Instruct     & \ding{55} & \ding{55} & \ding{55} & \ding{55} & 5.59 \\
    Qwen 2.5-VL 7B Instruct     & \ding{55} & \ding{55} & \ding{55} & \ding{55} & 5.40 \\
    Ego-status MLP & \ding{55} & \ding{55} & \ding{55} & \ding{55} & 7.47 \\
    \midrule
    Poutine-Base CoVLA          & \ding{51} & \ding{55} & \ding{51} & \ding{55} & 7.74 \\
    Poutine-Base No-CoVLA       & \ding{55} & \ding{51} & \ding{51} & \ding{55} & 7.95 \\
    Poutine-Base No Language    & \ding{51} & \ding{51} & \ding{55} & \ding{55} & 7.94 \\
    Poutine-Base                & \ding{51} & \ding{51} & \ding{51} & \ding{55} & \textbf{8.12} \\
    Poutine-Base CoT            & \ding{51} & \ding{51} & \ding{51} & \ding{51} & \underline{8.08} \\
    \midrule
    \color{gray}Ground-Truth    & \color{gray}– & \color{gray}– & \color{gray}– & \color{gray}– & \color{gray}8.13 \\
    \bottomrule
  \end{tabular}
  \caption{\textbf{Results on WOD-E2E validation split with different pre-training and inference strategies.} Best-performing model \textbf{bolded} and second-best \underline{underlined}.}
  \label{tab:pre-training-results}
\end{table*}

\textbf{Datasets.} For VLT pre-training, we leveraged the CoVLA~\cite{arai2025covla} and WOD-E2E datasets. CoVLA contains 10,000 front-view, 30~s driving videos recorded in Japan at 20 Hz with corresponding ego trajectory information. For each frame, we stored the future 5~s ego trajectory subsampled to 4 Hz to be consistent with the WOD-E2E format. The WOD-E2E dataset provides 4021 long-tail driving scenarios of 20 s each, captured at 10 Hz from 8 multi-view cameras. Each scenario belongs to one of 10 (11 for test set) predefined long-tail scenario categories, such as pedestrians, intersections, cut-ins, and special vehicles, where the degree of `rarity' significantly varies across scenarios. The official split allocates 2037 videos for training, 479 for validation, and 1505 for testing. Each frame contains a 4~s ego-vehicle past trajectory and a 5~s ground-truth future trajectory, both sampled at 4 Hz. Only the three front-facing camera images from the current frame in the WOD-E2E dataset were used as input to the model\footnote{Although omitting side and rear views reduced contextual coverage, we found that the front images are sufficient for most scenarios and their smaller token count enabled substantially more optimization steps within our compute budget.}, whereas we used three consecutive frames from the preceding second for CoVLA (see Fig.~\ref{fig:intro-fig} for an example). The raw images were downsampled to a maximum resolution of $H \times W = 512\times 512$ pixels from their original sizes of $1208\times 1928$ (CoVLA) and $1079 \times 972$ pixels (WOD-E2E). The WOD-E2E validation set contains 479 human preference-labeled frames, used as part of our RL fine-tuning stage.

Language annotations were automatically generated using our proposed labelling method from Section \ref{sec:ann}. We generated 240.5~K high-quality captions sampled at 1 Hz for the WOD-E2E (training set) and CoVLA datasets\footnote{The original CoVLA captions lacked semantic diversity which prompted us to generate new captions.}, using 12 A100 GPUs over 24 h. To accelerate VLM inference, front-view images from the current frame were resized to 532 × 476 px, while all other images were limited to at most 256 × 256 px. Table \ref{tab:ann-dataset-stats} summarizes key details of our language‑annotation dataset. To fully exploit the available data for training, the training set was subsampled so that captions accompanied half of the selected frames. In total, we used $\approx$10\% of CoVLA frames and 20\% of WOD-E2E frames for model training, uniformly subsampled across scenarios.
\\\\
\textbf{Evaluation Metrics.} Average displacement error (ADE) and rater-feedback score (RFS) were the two metrics used in our open-loop evaluations on the WOD-E2E benchmark. ADE is computed as the L2 distance between the predicted and reference trajectories, averaged over all time steps up to the specified horizon. Here, the reference trajectory for the ADE is the highest rater-scored trajectory, described below. Each frame with preference labels in the WOD-E2E dataset includes three 5-second trajectories rated by human experts on a scale from 0 to 10, with 10 indicating the most preferred driving behavior. Computing RFS requires identifying the rater-scored trajectory that best aligns with the model prediction. Predictions that lie within the trust region (at 3 s and 5 s) of its nearest rated trajectory get assigned that particular preference score. Otherwise, the assigned score is exponentially lower than that of the nearest rated trajectory's score, with a floor RFS of 4. The overall RFS score—used as the official ranking metric on the WOD-E2E challenge—is computed as the mean of the average RFS scores across scenario categories.
\\\\
\textbf{Implementation Details.} The training of Poutine-Base was completed in two stages: VLT pre-training on CoVLA followed by VLT pre-training on the WOD-E2E training data. For CoVLA VLT pre-training, we fine-tuned all modules of the \texttt{Qwen-2.5-VL-3B-Instruct} model for one epoch with an effective batch size of 64 and a learning rate of 1e-5 under a cosine decay schedule, completing in 24 h on four NVIDIA A100 GPUs. WOD-E2E VLT pre-training used identical hyper-parameters except for two epochs and a reduced batch size of 16, finishing in 10 h on the same hardware. RL post-training then optimized Poutine-base with GRPO for 2,000 steps using the preference-labeled trajectories provided in the WOD-E2E validation data along with their corresponding rater feedback labels to compute the RFS (normalized to 1) as the reward signal. We employed a linear-decay schedule from 1e-6, sampling temperature 0.9, $\beta=0.04$, 8 rollouts per sample, and an effective batch size of 32. This stage required 12 h on four A100 GPUs. At inference, we used a 1e-6 temperature for the CoT and greedy decoding for trajectory prediction. We removed the navigation command from conditioning at inference, which we found to marginally improve performance.

\noindent \textbf{Results.}  We present results from evaluating our approach on the WOD-E2E benchmark. Table~\ref{tab:waymo-test} shows all submissions\footnote{All submissions as of \textit{June 12, 2025}. Only the top-performing submission (by overall RFS) is shown for each method.} that outperform the Waymo baseline on the WOD-E2E test set, listed in ascending order of the overall RFS score. A small MLP of size 300 K parameters trained solely on ego status information (i.e., no vision information) is also shown as a baseline. Our \textit{Poutine} model tops the leaderboard by a wide margin with an RFS of 7.99, while its pre-RL variant, \textit{Poutine-Base}, achieves a score of 7.91 (second overall on the leaderboard) highlighting the value of GRPO post-training. Among the top submissions, \textit{Poutine} also achieves the best ADE between the highest rater-scored trajectory and the model prediction at 3 s and 5 s. 

Although the RL fine-tuning stage improves both overall RFS and ADE compared to \textit{Poutine-Base}, we observe that the performance drops in several scenario categories. Specifically, \textit{Poutine} underperforms \textit{Poutine-Base} in Spotlight, Construction, and Multi-lane maneuvers—three out of the 11 total scenario categories. Interestingly, \textit{Poutine-Base} achieves the best overall RFS score in Spotlight, which consists of manually selected, particularly challenging scenarios. More work is required to better understand the impact of fine-tuning on preference data and how it affects the overall driving policy, particularly on rarer scenarios.

Note that while AutoVLA \cite{zhou2025autovla} shares significant overlap with Poutine, it demonstrates a limited improvement (+0.03 RFS) compared to Poutine's substantial gain (+0.46 RFS) relative to the Waymo baseline. One possible explanation is that the driving-specific action tokenizer used by AutoVLA limits the ability to leverage world knowledge embedded in the base VLM, compared to directly encoding
trajectories in text space.
\\\\
\textbf{Ablations.}
Several ablations were conducted to assess the various design decisions of Poutine using the WOD-E2E validation set. Table \ref{tab:pre-training-results} analyzes pre-training ablations. VLMs that skip VLT pre-training (i.e., the two Qwen baselines) trail all Poutine variants, confirming the necessity of domain-specific pre-training. Training on CoVLA alone (\textit{Poutine-Base CoVLA}) transfers zero-shot to Waymo with an RFS of 7.74, demonstrating strong cross-domain generalization. Removing CoVLA data (\textit{Poutine-Base No-CoVLA}) or the auto-generated captions (\textit{Poutine-Base No Language}) from training degrades performance to 7.95 and 7.94, respectively, showing that both large-scale pre-training and language supervision are beneficial. Consistent with prior works \citep{hwang2024emma, sima2024drivelm}, generating CoT at inference (\textit{Poutine-Base CoT}) does not improve over the no-CoT \textit{Poutine-Base} variant. Further investigation is required to determine the benefit of CoT for reasoning in long-tail scenarios at inference time. The final \textit{Poutine-Base} model achieves an 8.12 RFS, nearly matching the 8.13 RFS attained with the expert WOD-E2E trajectories.

\begin{figure}
    \centering
    \includegraphics[width=\columnwidth]{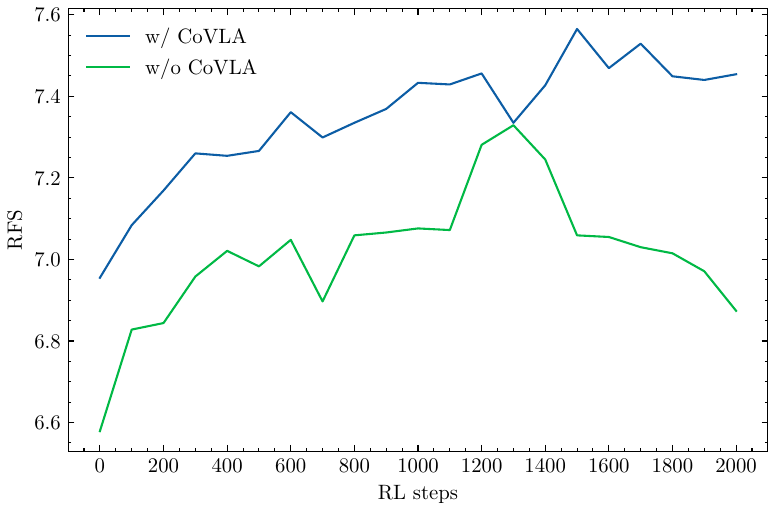}
    \caption{\textbf{GRPO Results.} Comparison between RL on a model pretrained with (blue) versus without (green) CoVLA. Both models were pre-trained on WOD-E2E training data. Checkpoints were evaluated on a held-out test set of 63 examples from the WOD-E2E validation split. }
    \label{fig:rl_covla_vs_no_covla}
\end{figure}

Figure \ref{fig:rl_covla_vs_no_covla} shows the results of applying GRPO to the models pre-trained with and without CoVLA data for 2,000 steps on 416 of the 479 WOD-E2E preference-labeled validation examples, with the remaining 63 held-out for evaluation purposes. Both models clearly benefit from RL post-training; however, the model pre-trained with CoVLA achieves significantly higher RFS, illustrating that large-scale VLT pre-training provides a stronger foundation for subsequent RL optimization.

%% file: sec/5_conclusion.tex
\section{Conclusion} Poutine combines VLT pre-training with GRPO fine-tuning to yield a 3B-parameter VLM that sets a new state-of-the-art on the Waymo Vision-Based E2E Driving benchmark. Its simple and scalable design highlights a practical path towards robust long-tail driving autonomy.

Future work includes extending our evaluations to closed-loop benchmarks, better understanding the impact of RL post-training, and determining an effective strategy to do CoT reasoning at inference time such that driving performance substantially improves on out-of-distribution scenarios. An interesting direction is performing reasoning in latent space, which may offer a more natural
representation to capture the uncertain and multi-modal nature of the reasoning trace compared to
language.

\section*{Acknowledgment}
We thank Samsung, the IVADO and the Canada First Research Excellence Fund (CFREF) / Apogée Funds, the Canada CIFAR AI Chairs Program, and the NSERC Discovery Grants program for financial support. We also thank Mila - Quebec AI Institute for compute resources.

%% file: sec/X_suppl.tex
\clearpage
\setcounter{page}{1}

\appendix

\section{Appendix}

\subsection{System prompts}

\label{app:annotation}
We include the system prompts used by the VLMs: Figure \ref{fig:system-prompt-ann} shows the prompt for generating language annotations, while Figure \ref{fig:system-prompt-sft} presents the prompts employed during supervised fine‑tuning and RL post-training.

\begin{figure}[h]
  \centering
  \begin{minipage}{\linewidth}
    \lstset{style=prompt}
    \begin{lstlisting}
You are an expert labeller of driving scenarios.
Input:
- 3 frames of multi-view images collected from the ego-vehicle over the last 1 second
- Current high-level intent (string)
- 4-second past trajectory (16 steps at 4 Hz)
- Expert 5-second future trajectory (20 steps at 4 Hz)
Task:
1. Inspect the input and decide, for each object class below, whether at least one critical instance of that class is present (i.e., it materially affects the ego-vehicle's future trajectory). A vehicle can be a car, bus, truck, motorcyclist, scooter, etc. traffic_element includes traffic signs and traffic lights. road_hazard may include hazardous road conditions, road debris, obstacles, etc. A conflicting_vehicle is a vehicle that may potentially conflict with the ego's future path.
   Object classes to audit:
     - nearby_vehicle
     - pedestrian
     - cyclist
     - construction
     - traffic_element
     - weather_condition
     - road_hazard
     - emergency_vehicle
     - animal
     - special_vehicle
     - conflicting_vehicle
     - door_opening_vehicle
2. Output "yes" or "no" for every class (no omissions).
3. Compose a concise natural-language description explaining why the expert safe driver plans the given future trajectory.
     - Mention only the classes you marked "yes"
     - Describe how each of those critical objects or conditions influences the trajectory.
     - Do not invent objects or conditions not present in the input.
4. From the expert's 5-second future trajectory, assign exactly one category from each list:  
     - speed $\in$ { keep, accelerate, decelerate }  
     - command $\in$ { straight, yield, left_turn, right_turn, lane_follow, lane_change_left, lane_change_right, reverse }  
   Choose the label that best summarises the overall behaviour of the expert future trajectory.  
     - If none fits, use `other`, but do this sparingly.
Output format (strict JSON, no extra keys, no commentary):
{
  "critical_objects": {
    "nearby_vehicle": "yes | no",
    "pedestrian": "yes | no",
    "cyclist": "yes | no",
    "construction": "yes | no",
    "traffic_element": "yes | no",
    "weather_condition": "yes | no",
    "road_hazard": "yes | no",
    "emergency_vehicle": "yes | no",
    "animal": "yes | no",
    "special_vehicle": "yes | no",
    "conflicting_vehicle": "yes | no",
    "door_opening_vehicle": "yes | no"
  },
  "explanation": "100-word description that references only the classes marked 'yes'",
  "meta_behaviour": {
    "speed": "keep | accelerate | decelerate | other",
    "command": "straight | yield | left_turn | right_turn | lane_follow | lane_change_left | lane_change_right | reverse | other"}}
    \end{lstlisting}
  \end{minipage}
  \caption[System prompt for language-annotation generation]{%
    \textbf{System prompt for generating language annotations.} We changed `driving scenarios' to `\emph{left-hand-side} driving scenarios' and `multi-view images' to `\emph{front-view} images' for CoVLA. 
  }
  \label{fig:system-prompt-ann}
\end{figure}

\begin{figure}[t]
  \centering
  \begin{minipage}{\linewidth}
    \lstset{style=prompt}
    \begin{lstlisting}
You are an expert driver.
Input:
- 1 frame of multi-view images collected from the ego-vehicle at the present timestep
- Current high-level intent (string)
- 4-second past trajectory (16 steps at 4 Hz)
Task 1: Critical Objects and Conditions Detection
Decide whether at least one critical instance of each class could influence the ego-vehicle's future path (no omissions). A vehicle can be a car, bus, truck, motorcyclist, scooter, etc. traffic_element includes traffic signs and traffic lights. road_hazard may include hazardous road conditions, road debris, obstacles, etc. A conflicting_vehicle is a vehicle that may potentially conflict with the ego's future path. Output "yes" or "no" for every class (no omissions).
   Object classes to audit:
     - nearby_vehicle
     - pedestrian
     - cyclist
     - construction
     - traffic_element
     - weather_condition
     - road_hazard
     - emergency_vehicle
     - animal
     - special_vehicle
     - conflicting_vehicle
     - door_opening_vehicle
Output format (strict JSON, no extra keys, no commentary):
{
  "critical_objects": {
    "nearby_vehicle": "yes | no",
    "pedestrian": "yes | no",
    "cyclist": "yes | no",
    "construction": "yes | no",
    "traffic_element": "yes | no",
    "weather_condition": "yes | no",
    "road_hazard": "yes | no",
    "emergency_vehicle": "yes | no",
    "animal": "yes | no",
    "special_vehicle": "yes | no",
    "conflicting_vehicle": "yes | no",
    "door_opening_vehicle": "yes | no"
  }}
Task 2: Natural Language Explanation
Compose a concise natural-language description of the optimal future 5-second trajectory for the ego vehicle that the expert driver (you) plans and explain why the expert driver plans to execute this trajectory.
    - Mention only the classes you marked "yes" in the previous task.
    - Describe how each of those critical objects or conditions influences the optimal trajectory.
    - Do not invent objects or conditions not present in the input.
Output format (strict JSON, no extra keys, no commentary):
{
  "explanation": "100-word description that references only the classes marked 'yes'"
}
Task 3: Meta-Behaviour Selection
Assign exactly one category from each list. Choose the label that best summarises the overall behaviour of the optimal future trajectory:  
     - speed $\in$ { keep, accelerate, decelerate }  
     - command $\in$ { straight, yield, left_turn, right_turn, lane_follow, lane_change_left, lane_change_right, reverse }  
     - If none fits, use `other`, but do this sparingly.
Output format (strict JSON, no extra keys, no commentary):
{
  "meta_behaviour": {
    "speed": "keep | accelerate | decelerate | other",
    "command": "straight | yield | left_turn | right_turn | lane_follow | lane_change_left | lane_change_right | reverse | other"
  }}
Task 4: Future Trajectory Prediction
Given the input, critical objects/conditions, natural language explanation, and meta-behaviour, predict the optimal 5-second future trajectory (5 steps at 1 Hz) of the ego vehicle.
Output format (raw text, not markdown or LaTeX):
[x_1, y_1], [x_2, y_2], [x_3, y_3], [x_4, y_4], [x_5, y_5]
    \end{lstlisting}
  \end{minipage}
  \begin{minipage}{\linewidth}
    \lstset{style=prompt}
    \begin{lstlisting}
You are an expert driver.
Input:
- 1 frame of multi-view images collected from the ego-vehicle at the present timestep
- Current high-level intent (string)
- 4-second past trajectory (16 steps at 4 Hz)
Task: Future Trajectory Prediction
1. Given the input, predict the optimal 5-second future trajectory (5 steps at 1 Hz) of the ego vehicle.
Output format (raw text, not markdown or LaTeX):
Future trajectory: [x_1, y_1], [x_2, y_2], [x_3, y_3], [x_4, y_4], [x_5, y_5]
    \end{lstlisting}
  \end{minipage}
  \caption[System prompt for VLT pre-training]{%
    \textbf{System prompt used for VLT pre-training.} (Top) Prompt used for frames with language annotations. (Bottom) Prompt used for frames without language annotations. This is also the prompt used during the RL post-training stage. The prompts were adjusted accordingly for pre-training on CoVLA data. %
  }
  \label{fig:system-prompt-sft}
\end{figure}